\documentclass[conference]{IEEEtran}
\IEEEoverridecommandlockouts
\usepackage{cite}
\usepackage{amsmath,amssymb,amsfonts}
\usepackage{algorithmic}
\usepackage{graphicx}
\usepackage{textcomp}
\usepackage{xcolor}
\usepackage{algorithm}
\usepackage{subfigure}
\usepackage{float}
\usepackage[backref]{hyperref}
\hypersetup{hidelinks,
	colorlinks=true,
	allcolors=black,
	pdfstartview=Fit,
	breaklinks=true}

\def\BibTeX{{\rm B\kern-.05em{\sc i\kern-.025em b}\kern-.08em
    T\kern-.1667em\lower.7ex\hbox{E}\kern-.125emX}}

\usepackage[
left=0.62in,right=0.62in,top=0.75in,bottom=1.0in]{geometry}

\newcommand{\nop}[1]{} 

\begin{document}

\title{Traffic Matrix Estimation based on Denoising Diffusion Probabilistic Model}

\author{\IEEEauthorblockN{Xinyu Yuan\IEEEauthorrefmark{2}, Yan Qiao\IEEEauthorrefmark{2}\IEEEauthorrefmark{1}\thanks{\IEEEauthorrefmark{1} Yan Qiao is the corresponding author.}, Pei Zhao\IEEEauthorrefmark{2}, Rongyao Hu\IEEEauthorrefmark{2}, Benchu Zhang\IEEEauthorrefmark{2}}
\IEEEauthorblockA{\IEEEauthorrefmark{2} School of Computer Science and Information Engineering, Hefei University of Technology, Hefei, China}
\IEEEauthorblockA{Emails: \href{qiaoyan@hfut.edu.cn}{qiaoyan@hfut.edu.cn\IEEEauthorrefmark{1}}, \href{yxy5315@gmail.com}{yxy5315@gmail.com}}
}

\maketitle


\begin{abstract}
The traffic matrix estimation (TME) problem has been widely researched for decades of years. Recent progresses in deep generative models offer new opportunities to tackle TME problems in a more advanced way. In this paper, we leverage the powerful ability of denoising diffusion probabilistic models (DDPMs) on distribution learning, and for the first time adopt DDPM to address the TME problem. To ensure a good performance of DDPM on learning the distributions of TMs, we design a preprocessing module to reduce the dimensions of TMs while keeping the data variety of each OD flow. To improve the estimation accuracy, we parameterize the noise factors in DDPM and transform the TME problem into a gradient-descent optimization problem. Finally, we compared our method with the state-of-the-art TME methods using two real-world TM datasets, the experimental results strongly demonstrate the superiority of our method on both TM synthesis and TM estimation.

\end{abstract}

\begin{IEEEkeywords}
denoising diffusion probabilistic models, deep learning, traffic matrix estimation, network tomography
\end{IEEEkeywords}

\section{Introduction}\label{1}


The traffic matrix (TM) represents the traffic demands between all possible pairs of network nodes, which are usually mentioned as origin to destination (OD) flows~\cite{r3}. It can provide critical knowledge of traffic status for solving many network management problems, such as traffic engineering, anomaly detection, and capacity planning~\cite{r29}. However, with the continuous expansion of network scale, measuring all OD flows directly by collecting each packet trace requires extremely high administrative costs as well as computational overhead~\cite{r15}\cite{r31}\cite{r1}. A more practical way is to estimate the flow-level traffic from the link-level traffic by means of network tomography (NT)~\cite{r7}.

NT-based traffic matrix estimation (TME) methods infer the fine-grained OD flows by solving a group of linear equations that involve both coarse-grained link loads and flow routing matrix. However, the key problem for NT-based TME is the linear equations are usually highly rank deficient, which means there is no unique solution of OD flows corresponding to the measured link loads. To tackle the problem, numerous types of research, and decades of years, explore the pattern of OD flows and transform the ill-posed inverse problem into a constrained inverse problem. So that the variables of OD flows may have a unique solution. For example, Vardi et al.~\cite{r6} and Tebaldi et al.~\cite{r31} suppose the volumes of OD flows satisfy a Poisson traffic model, while Cao et al.~\cite{r32} assume a Gaussian traffic model. Zhang et al.~\cite{r24} assume that the OD flows follow an underlying gravity model. However, as the performance of the above methods highly relies on the accuracy of these assumptions, there is a lack of generality.  

With the rapid progress in deep learning technologies, deep generative models, such as Variational Autoencoder (VAE)~\cite{r20} and Generative Adversarial Network (GAN)~\cite{r30}, appear as powerful pattern learning tools. By training a generative model with a group of samples, the model can automatically learn the training data patterns and generate synthetic samples with almost the same distribution as the training ones. Two recent works \cite{r11} and \cite{r12} respectively utilize VAE and GAN to tackle the TME problem. The key idea is to first train VAE/GAN with a group of historical TMs, then use the trained model to generate the estimated TM most consistent with the tomography equations. The results in \cite{r11} and \cite{r12} showed a significant improvement over past assumption-based methods. However, either VAE or GAN has its inherent model defects: VAE tends to produce unrealistic, blurry samples while the training of GAN is often unstable. Moreover, as they generate synthetic samples from a Gaussian noise through ``one-step'' transformation, they hardly work well on generating samples with complex distributions (such as TMs).

The denoising diffusion probabilistic model (DDPM) is a recently proposed generative model~\cite{r13}, which has shown its superiority over VAE and GAN in generating samples from complex data distributions. The superior performance of DDPM is owning to the ``multi-step'' processes that transform noise into a synthetic sample following the Markov chain of diffusion. Unfortunately, applying DDPM directly to deal with the TME problem is intractable. There are mainly three barriers: First, as the transformation from noises to synthetic instances usually involves thousands of steps, the computing time is extremely long when TM is on a large scale. Second, since the magnitude differences between OD flows are generally large, the common normalization on TM will prevent DDPM from learning the true distributions. Last but not least, unlike VAE or GANs sampling from a single low-dimensional vector, diffusion models are learned with a fixed procedure from latent space, in which the latent vectors have the same high dimensions as the original data, with numerous independent noises involved in each sampling step.

In this paper, we make the first attempt to develop a DDPM-based TME approach (named DDPM-TME) that can accurately solve the ill-posed inverse problem even when the underlying traffic is within a complex distribution. The new approach not only overcomes the three barriers mentioned above but also demonstrates that DDPM can help to generate synthetic TMs that are surprisingly close to the real ones.

In Summary, the main contributions of this paper can be summarized as follows.

\begin{itemize}
\item We propose a DDPM-based TME model that can accurately estimate all OD flows in TM by inputting the link loads. Thanks to the powerful ability of DDPM on learning complex distributions, the performance of our model has significant superiority over VAE and GAN. The new model can not only accurately estimate the TM from link loads, but also can be used as a TM simulator for the scenario where true TMs of the underlying network are not currently available.

\item We design a preprocess module before the DDPM network to improve the performance of distribution learning. The preprocessing module provides a reversible mapping between real-TM and latent representations. On one hand, it can reduce the high dimensionality of the DDPM learning space. On the other, it scales the flows in TM to a much smaller range space without losing the dynamics of the distribution.

\item We design a gradient-descent optimization method that can estimate TMs accurately with the trained model. Rather than solving constrained linear equations direct, we parameterize the noises in each sampling step of DDPM and transform the ill-posed inverse problem into a gradient-descent optimization problem. To accelerate the sampling process, we adopted a fast version of diffusion that can greatly cut down the sampling time of DDPM. 


\item We compared our method with the state-of-the-art TME methods that are based on deep generative models, using two real-world traffic datasets. The experimental results showed that our method can improve the estimation accuracy of the VAE-based method by $68\%$ and the GAN-based method by $25\%$. Furthermore, the synthetic TMs outputted by our model are much closer to the real TMs than other generative models. We published the source code under the MIT license along with detailed documents at \href{https://github.com/Y-debug-sys/DDPM-TME}{\textcolor{magenta}{https://github.com/Y-debug-sys/DDPM-TME}}.
\end{itemize}

\section{Problem Formulation And Background}\label{3}

\subsection{TM Estimation Problem Formulation}\label{3a}
We model the network graph as $G=(V, E)$ where $V$ and $E$ are network nodes and links, respectively. The TM is defined as a $|V|\times |V|$ matrix where each entry represents an OD flow between a pair of nodes in the network. To facilitate calculations, we reshape TM to a vector $X=\{x_1,x_2,\dots,x_n\}$, where $n=|V|\times|V|$ is the total number of OD flows in TM. We denote a sequence of TMs from time point $1$ to $T$ as $X=\left\{X(1),X(2),\dots,X(T)\right\}$, where $X(t)=\{x_{1}(t),x_{2}(t),\dots,x_{n}(t)\}$ denotes the TM at time point $t\in T$. Let $Y=\left\{Y(1),Y(2),\dots,Y(T)\right\}$ denote the sequence of link loads, and $A\in \mathbb{R}^{m\times n}$ denote the routing matrix, where each entry $a_{ij}$ of $A$ has a binary value ($0$ or $1$). For deterministic routing policy, if the $j$-th flow traverses the $i$-th link, then $a_{ij}=1$; otherwise, $a_{ij}=0$. For probabilistic routing policy (such as ECMP), the value of $a_{ij}$ is within the range of $[0,1]$, representing the probability that the $j$-th flow may transverse the $i$-th link. The relationship between TM $X$ and link load $Y$ can be formulated as the linear equations:
\begin{equation}
AX=Y\label{eq1}
\end{equation}

For most networks, since the number of OD flows is always larger than the number of link loads, i.e., $n\gg m$, the problem of TM estimation is under-determined. That means Eqn.~\eqref{eq1} does not have a unique solution in most cases and cannot be directly solved without additional constraints on $X$.

\subsection{Denoising Diffusion Probabilistic Models}\label{3b}

\begin{figure}
\centerline{\includegraphics[width=1\linewidth]{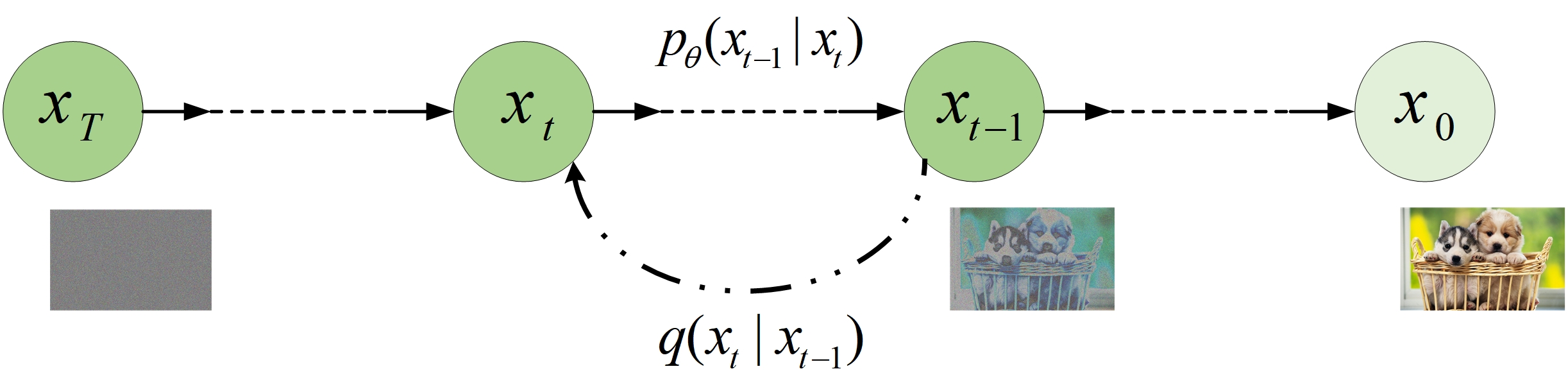}}
\caption{Graphical models for diffusion in DDPM.}
\label{fig_1}
\end{figure}

Denoising diffusion probabilistic model (DDPM) \cite{r13} proposed recently has been widely recognized as the state-of-the-art generative model with respect to generating high-quality samples \cite{r18}. The classical DDPM is built on two diffusion processes: a forward diffusion known as the noising process that drives any data distribution to a tractable distribution such as Gaussian by gradually adding noise to the data, and a backward diffusion known as the denoising process that sequentially removes noise from noised data to produce synthetic samples.

In particular, DDPMs (as shown in fig.~\ref{fig_1}) are latent variable models of the form
\begin{equation}
p_{\theta}(x_0)=\int p_{\theta}(x_{0:T})dx_{1:T}\label{eq2}
\end{equation}
where $p_{\theta}(x_{0:T}):=p_{\theta}(x_T)\prod_{t=1}^{T}p_{\theta}^{(t)}(x_{t-1}|x_{t})$, and $x_1,\dots,x_T$ are latent variables in the same sample space as $x_0$. The parameters $\theta$ are learned to fit the data distribution $q(x_0)$ by maximizing a variational lower bound:
\begin{equation}
\mathbb{E}_{q}\left[{-\log{p_\theta }({x_0})}\right]\le
\mathbb{E}_{q}[-\log p({x_T})-\sum\limits_{t\ge1}{\log\frac{{{p_\theta}({x_{t-1}}|{x_t})}}{{q({x_t}|{x_{t-1}})}}}]\label{eq3}
\end{equation}

What distinguishes diffusion models from other types of latent variable models is the approximate posterior $q(x_{1:T}|x_0)$ (called the forward process) is fixed to a Markov chain according to a variance schedule $\beta_1,\dots,\beta_T$:
\begin{equation}
q({x_{1:T}}|{x_0}): = \prod\limits_{t = 1}^T {q({x_t}|{x_{t - 1}})}\label{eq4}
\end{equation}
where $q({x_t}|{x_{t - 1}}):={\cal N}\left({{x_t};\sqrt {1-{\beta _t}}{x_{t-1}},{\beta _t}I}\right)$. For $t=1,\dots,T$, the noise at each step is defined by ${\beta _t}\in\left({0,1}\right)$. It can be formulated as a small linear schedule and was updated by the schedule method in \cite{r13} to increase linearly from ${\beta_1}={10^{-4}}$ to  
${\beta _T}=0.02$. 

A notable property of the forward process is that using notation $\alpha_t:=1-\beta_t$ and $\bar\alpha_t:=\prod\limits_{s=1}^t \alpha_s$, we can sample $x_t$ at any arbitrary time step $t$ in a closed form:
\begin{equation}
q({x_t}|{x_0}):=\int {q({x_{1:t}}|{x_0})d{x_{1:(t-1)}}}={\cal N}({x_t};\sqrt{{\bar\alpha _t}}{x_0},(1-{\bar\alpha_t})I)\label{eq5}
\end{equation}

Thus using reparameterization trick \cite{r20}, we have
\begin{equation}
x_t=\sqrt{{{\bar\alpha}_t}}x_0+\sqrt{1-{{\bar\alpha}_t}}\epsilon\label{eq6}
\end{equation}
where $\epsilon \sim{\cal N}(0,I)$. If all the conditionals are modeled as Gaussians with trainable mean functions and fixed variances, the objective in Eqn.~\eqref{eq3} can be simplified to:
\begin{equation}
{{\cal L}_{simple}}={\mathbb{E}_{t\sim[1-T],{x_0}\sim q({x_0}),\epsilon \sim {\cal N}(0,I)}}\left[{{{\left\|{\epsilon-{\epsilon _\theta}({x_t},t)}\right\|}^2}}\right]\label{eq7}
\end{equation}

Then the reverse process in DDPM can be written as:
\begin{equation}
{x_{t-1}}=\frac{1}{{\sqrt{{\alpha_t}}}}({x_t}-\frac{{1-{\alpha_t}}}{{\sqrt{1-{{\bar \alpha}_t}}}}{\epsilon_\theta }({x_t},t))+{\sigma_t}{z_{t-1}}\label{eq8}
\end{equation}
where $\sigma _t$ is hyperparameter fixed to $\sqrt{{\beta_t}}$, and $z_{t-1}$ is a standard Gaussian noise and $z_T=x_T$.

\section{Methodology} \label{4}
In this section, we develop a TME method to accurately estimate TMs from network link loads based on our new proposed DDPM-based model. The key idea is we first let our model learn the traffic pattern automatically from historical TMs, then we optimize the hyperparameters in the trained model to produce an optimal TM that most conforms to the input link loads.

\subsection{DDPM-TME model}\label{4a}

\begin{figure}[ht]
\centerline{\includegraphics[width=1\linewidth]{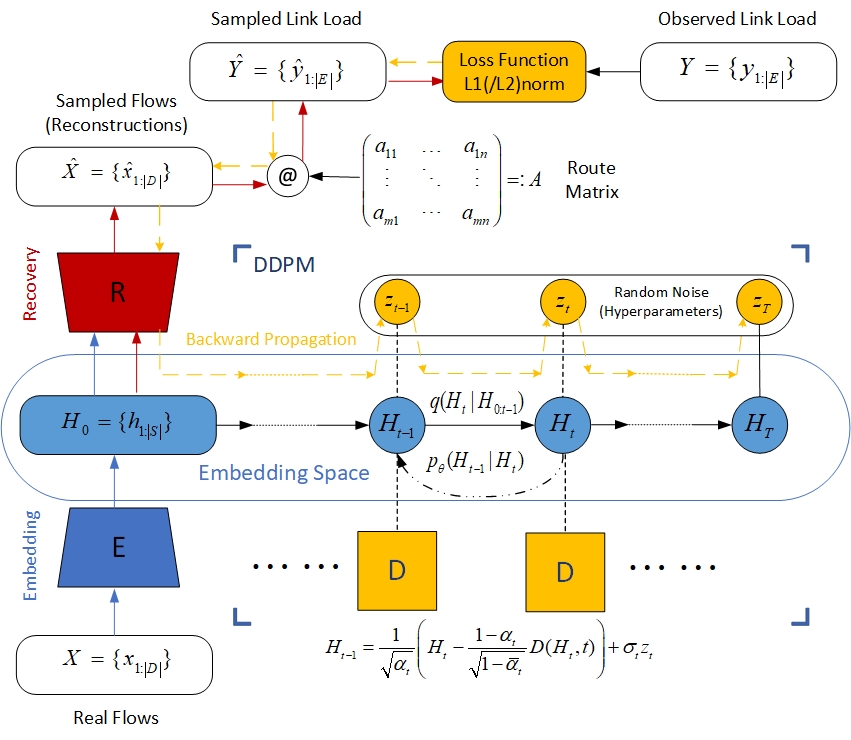}}
\caption{Overview of the proposed DDPM-based TME method.}
\label{fig2}
\end{figure}


Our DDPM-based model mainly contains two components, as shown in Fig.~\ref{fig2}: a preprocessing module and a DDPM network. The preprocessing module can encode the TM sample onto a latent space using an embedding network, and recover the noise in the latent space onto the sample space using a recovery network. Then DDPM network carries the forward and backward diffusion processes in the latent space rather than the sample space. Next, we explain why we add a preprocessing module before DDPM and how the model can be trained.

\subsubsection{Preprocessing module}
In the TME problem, the magnitude difference between OD flows can be extremely huge in practice, which may lead to oversized parameters in the model. In such a case, the most common way is to preprocess the training data with normalization or centralization to ensure the stability of training. For example, Min-Max normalization is one of the most general methods to scale the data on range $[0,1]$. However, existing normalization methods (such as Min-Max normalization) can not work well on TM data. That is because the volumes of OD flows are seriously polarized in some cases: for some time points, the vast majority of OD volumes are closed to zeros while very few flows appear as giant volumes. In such cases, Min-Max normalization will lead to distorted traffic distribution. 

Alternatively, representation learning, which was primarily used as an embedding method, can also be considered a data preprocessing tool. Note the fact that a complex system is usually controlled by the variations of some lower-dimensional factors. Hence, we design a data preprocessing module, that includes an embedding network and a recovery network, to encode the original TM onto a lower-dimensional latent space. So that the huge magnitude differences will be eliminated in the latent space without losing the key features of the traffic pattern. Allowing the DDPM to learn the traffic distribution via lower-dimensional representations, the embedding and recovery networks provide bidirectional mappings between feature and latent spaces. 

\subsubsection{Model Training}

\begin{algorithm}[htb] 
\caption{Training Algorithm of the DDPM-based Method} 
\label{alg1} 
\begin{algorithmic}[1] 
\REQUIRE 
$I_{max}$, $I_{pre}$, $\mathbf X=\{X(1),X(2),\dots,X(T)\}$
\ENSURE 
Trained $E$, $R$, $D$
\STATE $E, R, D \gets Initial()$; 
\label{1l1}
\STATE $i, j \gets 0$; 
\label{1l2}
\STATE $\mathbf {while}$ $i\le I_{pre}$ $\mathbf {do}$ 
\label{1l3}
\STATE ~~~ $\mathbf {for}$ $t\in \{1,2,\dots,T\}$ $\mathbf {do}$ 
\label{1l4}
\STATE ~~~~~~ Take gradient descent step on \\
~~~~~~~~~ $\nabla_{E,R}$ ${\Vert X(t)-R(E(X(t))) \Vert}^2$
\label{1l5}
\STATE ~~~~~~ $i \gets i+1$;
\label{1l6}
\STATE $\mathbf {while}$ $j\le I_{max}$ $\mathbf {do}$ 
\label{1l7}
\STATE ~~~ $\mathbf {for}$ $t\in \{1,2,\dots,T\}$ $\mathbf {do}$ 
\label{1l8}
\STATE ~~~~~~ Take gradient descent step on \\
~~~~~~~~~ $\nabla_{E,R}$ ${\Vert X(t)-R(E(X(t))) \Vert}^2$;
\label{1l9}
\STATE ~~~~~~ ${x_0} \leftarrow E(X(T\%(t+1)))$;
\label{1l10}
\STATE ~~~~~~ $t_s \gets Uniform(\{1,\dots,T_s\})$;
\label{1l11}
\STATE ~~~~~~ $\epsilon \sim{\cal N}(0,I)$;
\label{1l12}
\STATE ~~~~~~ Take gradient descent step on \\
~~~~~~~~~ $\nabla_{D}$ ${\Vert \epsilon-D(\sqrt {1-{{\bar \alpha }_{{t_s}}}}\epsilon +\sqrt{{{\bar \alpha}_{{t_s}}}}{x_0}, t_s) \Vert}^2$;
\label{1l13}
\STATE ~~~~~~ $j \gets j+1$;
\label{1l14}
\RETURN $E$, $R$, $D$; 
\end{algorithmic}
\end{algorithm}

To encourage networks to better understand the dynamics of the training data, instead of training the diffusion model with the fixed-trained embedding network, we jointly train the embedding and generator networks in each training epoch. The detailed training algorithm is shown in Alg.~\ref{alg1} where $t_s$ denotes timestep in a Markov chain, while $t$ presents the timepoint the TM sampled. Considering representations of original TMs cannot be extracted and restored well at the very beginning, the parameters in $E$ and $R$ will first be trained with $I_{pre}$ epochs, and then the joint training will last $I_{max}$ epochs until $D$ converged, given the training TM instances $\mathbf X=\{X(1), X(2),\dots, X(T)\}$.

\subsection{Traffic Matrix Estimation with DDPM-based Model} \label{4b}
Using the trained DDPM-based model, we aim to estimate an optimal TM from link loads using the powerful sample-producing ability of DDPM. As we have known, after training, the DDPM-based model can produce synthetic TMs that are quite ``similar'' to the realistic TMs. Hence, the TME problem in this section can be transformed into the problem that \textit{how to let the trained DDPM-based model produce an optimal synthetic TM that most conform with the input link loads?}


To answer the above question, a feasible way is to optimize the vector $z\in \mathbb{R}^k$ in the latent space using gradient descent to reduce the loss function $\left\|{\mathbf{A}G\left(z\right)-y}\right\|_2^2$, where $G(z)$ is the produced sample by the generative model. For example, the methods proposed in~\cite{r11} and~\cite{r12} respectively used VAE and GAN to generate the optimal TMs by optimizing the latent vector. However, such a method can not work well for DDPM. As mentioned in Section \ref{3b}, the production of a synthetic sample $x_0$ by DDPM needs to iteratively generate $x_t$ ($0\le t\le T-1$) in each step. That means the synthetic sample $x_0$ is not generated directly from the latent vector $x_T$. Hence, as $x_T$ is often hundreds or thousands of steps away from $x_0$, producing an optimal $x_0$ through optimizing the latent vector $x_T$ is not feasible.

To deal with the problem, we unfold the reverse process of DDPM (i.e., Eqn.~\eqref{eq8}) as follows:

\begin{equation}
{x_0}=\frac{1}{\sqrt{{\bar \alpha}_T}}x_T+\sum\limits_{t=1}^{T-1}\frac{\sigma_{t+1}}{\sqrt{{\bar \alpha}_t}}z_t
+F(z_{T},{z_{1:T-1}})\label{new1}
\end{equation}
where remainder term $F$ satisfies

\begin{equation}
F(x_{T},{z_{1:T-1}})=-\sum\limits_{t=1}^{T}\frac{1-\alpha_t}{\sqrt{({\bar \alpha}_t)(1-{\bar \alpha}_t)}}{{\epsilon _\theta}({x_t},t)}.\label{new2}
\end{equation}
Eqn.~\eqref{new1} and Eqn.~\eqref{new2} show that the impact of variables in each step is inversely proportional to the distance to $x_0$. It is also notable that noise $z_t$, $1\le t\le T-1$, also plays an important role in generating $x_0$. Therefore, to produce an optimal TM, we propose to view these random noises $z_t$, $1\le t\le T-1$, as hyperparameters, which will be optimized with latent vector $x_T$ together by gradient descent. Our experimental results demonstrate the proposed method works quite well on TME and achieved the best performance over former TME methods.

The overview of the proposed TME method based on the DDPM-based model is shown in Fig.~\ref{fig2}. As all training samples have been encoded onto the embedding space through the embedding network $E$, the DDPM network learns the inverse process of $q({H_t}|{H_{0:t-1}})$ in the embedding space and draws the samples from ${p_\theta}({H_{t-1}}|{H_t})$ in each step until $H_0$ is produced. Subsequently, $H_0$ is fed to the recovery network $R$ to recover the synthetic traffic $\hat X:=R({H_0})$. Hence, the problem of TM estimation can be formulated by:
\begin{equation}
\mathop {\arg \min}\limits_{z_{1:T}}\left\|{A \hat X(z_{1:T})-Y}\right\|^2\label{eq14}
\end{equation}
where $\left\| \cdot \right\|$ can be L1norm or L2norm. With the objective in Eqn.~\eqref{eq14}, we can compute the gradients of hyperparameters $z_{1:T}$ through backpropagation with a gradient descent optimizer.

As the DDPM network involves hundreds of diffusions for sampling, the proposed TME method may perform inefficiently in estimating large-scale TMs. To tackle this problem, we adopted DDIM \cite{r14}, a fast version of diffusion, to accelerate the running speed of our method. The DDIM has a similar training objective as DDPM but with a different sampling scheme:
\begin{equation}
\begin{aligned}
{x_{t-1}}=&\sqrt{{\bar\alpha_{t-1}}}\left({\frac{{{x_t}-\sqrt{1-{\bar\alpha_t}}{\epsilon_\theta}({x_t},t)}}{{\sqrt{{\bar\alpha_t}}}}}\right)+\\
&\sqrt{1-{\bar\alpha_{t-1}}-\sigma_t^2} \cdot {\epsilon_\theta}({x_t},t)+{\sigma_t}{\epsilon}\\
\end{aligned}\label{eq9}
\end{equation}
where ${\epsilon}\sim{\cal N}(0,I)$ is standard Gaussian noise that is independent of $x_t$. It uses a non-Markovian diffusion process leading to “shorter” generative Markov chains that can be simulated in a small number of steps.

\begin{algorithm}[htb] 
\caption{TME Algorithm Using DDPM-based Method} 
\label{alg2} 
\begin{algorithmic}[1] 
\REQUIRE 
$I_{opt}$, $N_{init}$, $D$, $R$, $\mathbf A$,\\
~~~~~~ $\mathbf Y=\{Y(1),Y(2),\dots,Y(B)\}$
\ENSURE 
Estimations $x(1),\dots,x(B)$
\STATE Generate random Gaussian noise set\\
~~ ${\epsilon_{set}}=\{\hat \epsilon(1), \dots,\hat \epsilon({N_{init}})\}$; 
\label{2l1}
\STATE $\{\hat x(1), \dots,\hat x({N_{init}})\} \gets BatchSample(D,R,\epsilon_{set})$; 
\label{2l2}
\STATE $i,j,k \gets 0$;
\label{2l3}
\STATE $\epsilon=\{\epsilon(1),\dots,\epsilon(B)\} \gets initial()$;
\label{2l4}
\STATE $\mathbf {while}$ $i\le B$ $\mathbf {do}$ 
\label{2l5}
\STATE ~~~ ${loss_{init}} \gets \infty$; 
\label{2l6}
\STATE ~~~ $\mathbf {while}$ $j\le N_{init}$ $\mathbf {do}$ 
\label{2l7}
\STATE ~~~~~~ $loss \gets {\Vert \mathbf A \hat x(j)-Y(i) \Vert}^2$;
\label{2l8}
\STATE ~~~~~~ $\mathbf {if}$ $loss\le loss_{init}$ $\mathbf {do}$
\label{2l9}
\STATE ~~~~~~~~~ $\epsilon(i) \gets \hat \epsilon(j)$, ${loss_{init}} \gets loss$;
\label{2l10}
\STATE ~~~~~~ $j \gets j+1$;
\label{2l11}
\STATE ~~~ $i \gets i+1$;
\label{2l12}
\STATE $\mathbf {while}$ $k\le I_{opt}$ $\mathbf {do}$ 
\label{2l13}
\STATE ~~~ $\{x(1), \dots, x(B)\} \gets BatchSample(D,R,\epsilon)$; 
\label{2l14}
\STATE ~~~ Take gradient descent step on \\
~~~~~~$\nabla_{\epsilon}($ ${\Vert \mathbf A \hat \{x(1), \dots, x(B)\}- \mathbf Y \Vert}^2)$;
\label{2l14}
\STATE ~~~ $k \gets k+1$;
\label{2l15}
\RETURN $x(1),\dots,x(B)$; 
\end{algorithmic}
\end{algorithm}


The TME algorithm using the DDPM-based model is presented in Alg.~\ref{alg2}. The algorithm inputs the routing matrix $\mathbf A$, the maximum number of optimization epochs $I_{opt}$, a batch of training instances $\mathbf Y$, “good” start point searching times $N_{init}$ as well as two trained models——recovery network $R$ and DDPM base model $D$. Along with some initial works (line 2$\sim$4), the algorithm first chooses a good starting point to further cut down the optimization steps. After generating $N_{init}$ groups of variables of noises $\epsilon_{set}$ (line 1), we select the best noise group, which can produce TM that most conform with the given link loads $\mathbf Y$ (line 5$\sim$11). Finally, Alg.~\ref{alg2} takes gradient descent optimization for $I_{opt}$ epochs (line 12$\sim$15).

\section{Experiments}\label{5}
In this section, we compare the performance of our DDPM-based TME method with VAE-based TME~\cite{r11} and GAN-based TME~\cite{r12}, respectively, on both TM synthesis and TM estimation.

\subsection{Datasets and Experimental Setup} \label{5a}
We evaluate the performance of our method using two real-world traffic datasets. The first dataset is from the Abilene network \cite{r24} which contains 12 routers. The TM samples in Abilene were collected every 5 minutes from March to September 2004. We use the TM samples during the first 16 weeks as the training data and the 17th week's samples as the testing data. The second dataset is from the GÉANT network \cite{r25} which has 23 routers. The TM samples were collected in the 15-minute interval from January to April 2004. We use the first 11 weeks’ collections as the training data and the $12$-th week’s collections as the testing data. 

In our experiments, we use a variant U-Net \cite{r26} architecture to build the DDPM network and two linear layers to build the embedding and recovery networks of the processing module in our model. The gradient descent optimization is carried out by the Adam optimizer~\cite{r33} with a learning rate of 1e-4. The detailed model hyperparameters are presented in Table~\ref{tab1}.

\vspace{-0.2cm}


\begin{table}[htbp]
\caption{Hyperparameters}
\begin{center}
\begin{tabular}{c | c}
\hline
FC layers & 2 (encoder), 1 (decoder)\\
\hline
Non-Linearity& LeakyReLu\\
\hline
Embedding dim& 128 (Abilene), 256 (GÉANT)\\
\hline
Decaying learning rate& $10^{-3} \sim 10^{-5}$\\
\hline
Inverse Multiquadratic& $1/(1 + \left\| {(x - x')/0.2} \right\|_2^2)$\\
\hline
Optimizer& Adam\\
\hline
Dropout& 0.0\\
\hline
Batch size& 32\\
\hline
Invertible blocks& 3\\
\hline
Training Epoch& 100\\
\hline
$\lambda_1$, $\lambda_2$, $\lambda_3$& 10, 2, 2\\
\hline
$\lambda_{zy}$, $\lambda_{x}$& 0.5, 1\\
\hline
\end{tabular}
\label{tab1}
\end{center}
\end{table}

Our method is implemented in PyTorch and all experiments were carried on a single NVIDIA Geforce RTX 3090 GPU. 

\subsection{Metrics} \label{5b}
To assess the quality of generated TMs, we apply t-SNE \cite{r27} and PCA \cite{r28} analyses on both original and synthetic TMs, which can visualize the similarity of their distributions in 2-dimensional space.

The performance of all TME methods is evaluated with three different metrics: the root mean square error ($RMSE$), the temporal-related mean absolute error ($TRE$), and the spatial-related mean absolute error ($SRE$).

Let $X(t) = \left\{{{x_1}(t),{x_2}(t),\dots,{x_n}(t)}\right\}$ denote the true values of OD flows measured at the $t$-th timepoint and $\hat X(t)=\left\{{{{\hat x}_1}(t),{{\hat x}_2}(t),\dots,{{\hat x}_n}(t)}\right\}$ denote the set of corresponding estimated values. The $RMSE$ can be calculated by:
\begin{equation}
RMSE(t)=\sqrt{\frac{1}{n}\sum\nolimits_{i = 1}^n{{{({x_t}(i)-{{\hat x}_t}(i))}^2}}} \label{eq16}
\end{equation}

The $TRE$ and $SRE$ can be calculated by:
\begin{equation}
TRE(t) = \frac{{\frac{1}{n}\sum\nolimits_{i=1}^n{\left|{{x_i}(t)-{{\hat x}_i}(t)}\right|}}}{{\frac{1}{n}\sum\nolimits_{i=1}^n{{x_i}(t)}}}\label{eq17}
\end{equation}
and
\begin{equation}
SRE(i) = \frac{{\frac{1}{{{T_{test}}}}\sum\nolimits_{t = 1}^{{T_{test}}}{\left|{{x_i}(t)-{{\hat x}_i}(t)}\right|}}}{{\frac{1}{{{T_{test}}}}\sum\nolimits_{t=1}^{{T_{test}}}{{x_i}(t)}}}\label{eq18}
\end{equation}
respectively.

\subsection{Baselines} \label{5c}
We compare our DDPM-based TME method with two recent generative-model-based TME methods: VAE \cite{r11} and WGAN-GP \cite{r12}, both of which used a trained decoder (or generator) for TM synthesis and estimation.

The model of VAE~\cite{r20} consists of an encoder in charge of mapping the traffics to vectors in latent space and a decoding network that maps the vectors back to the initial representations. Through a loss function minimizing both the Kulback-Leibler (KL) divergence and reconstruction difference, the VAE maximizes the probability of generating real data samples.

GAN~\cite{r30} is another classical generative model based on adversarial training. The WGAN improves the training performance of GAN by Wasserstein-1 distance and gradient penalty.
\vspace{\topsep}

\begin{figure}[htbp]
\centering
\subfigure[Our Model]{
\begin{minipage}[b]{.29\linewidth}
    \centering
    \includegraphics[width=1.1\linewidth]{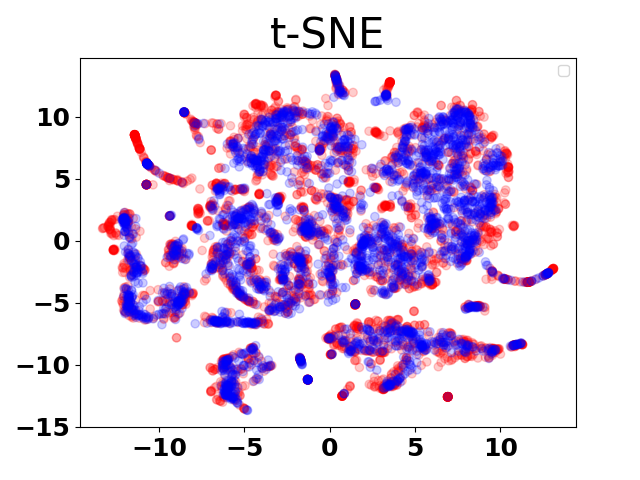} \\
    \includegraphics[width=1.1\linewidth]{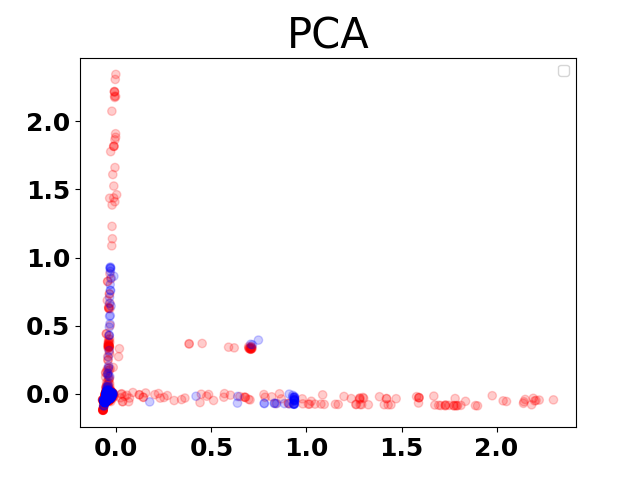}
\end{minipage}
}
\subfigure[VAE]{
\begin{minipage}[b]{.29\linewidth}
    \centering
    \includegraphics[width=1.1\linewidth]{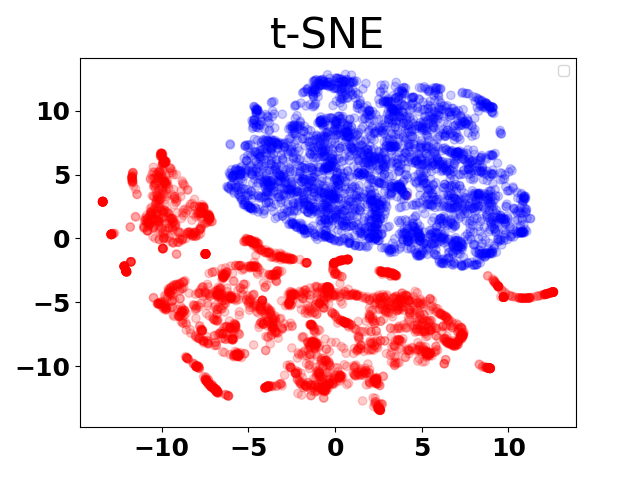} \\
    \includegraphics[width=1.1\linewidth]{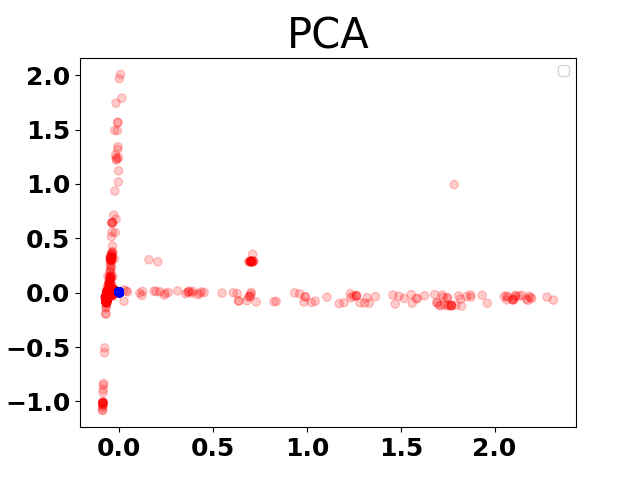}
\end{minipage}
}
\subfigure[WGAN]{
\begin{minipage}[b]{.29\linewidth}
    \centering
    \includegraphics[width=1.1\linewidth]{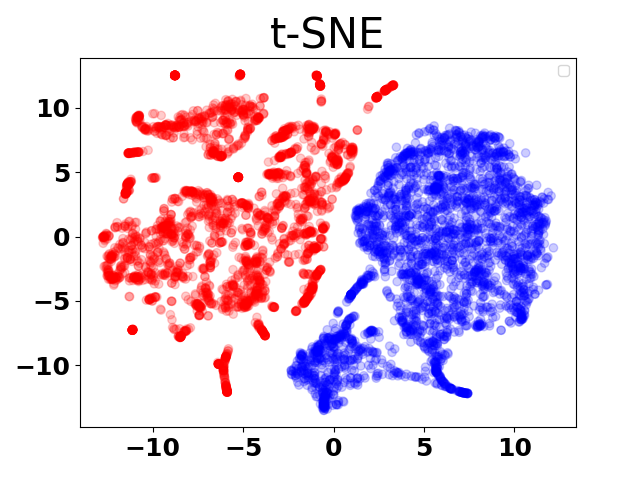} \\
    \includegraphics[width=1.1\linewidth]{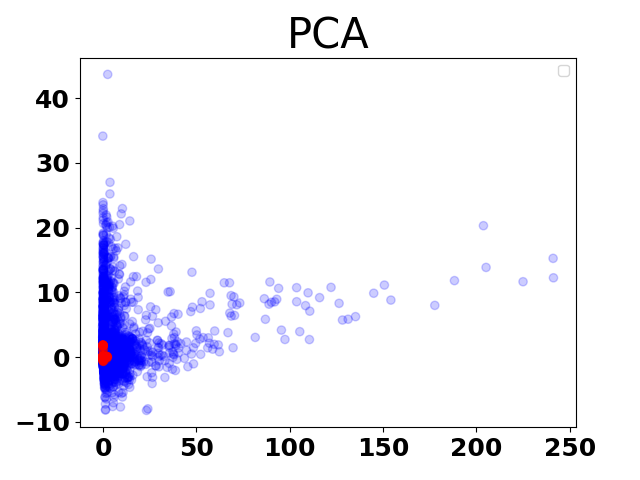}
\end{minipage}
}
\caption{The t-SNE and PCA plots of synthetic TMs in the Abilene dataset: the red dots denote the true TM data, and the blue dots denote the synthetic TM data.}
\label{fig_3}
\end{figure}
\vspace{-0.8cm}

\begin{figure}[htbp]
\centering
\subfigure[Our Model]{
\begin{minipage}[b]{.29\linewidth}
    \centering
    \includegraphics[width=1.1\linewidth]{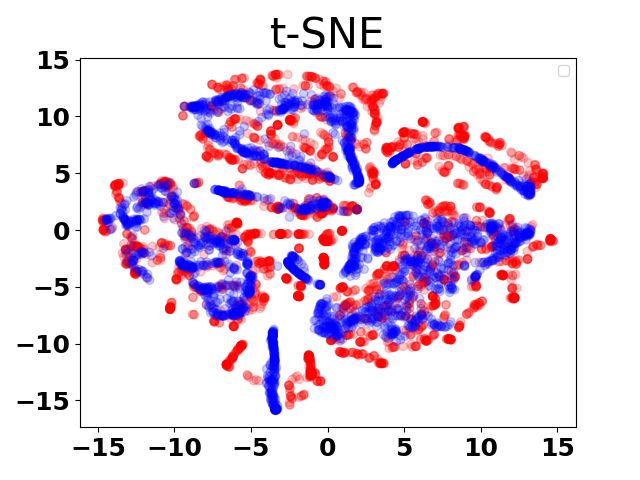} \\
    \includegraphics[width=1.1\linewidth]{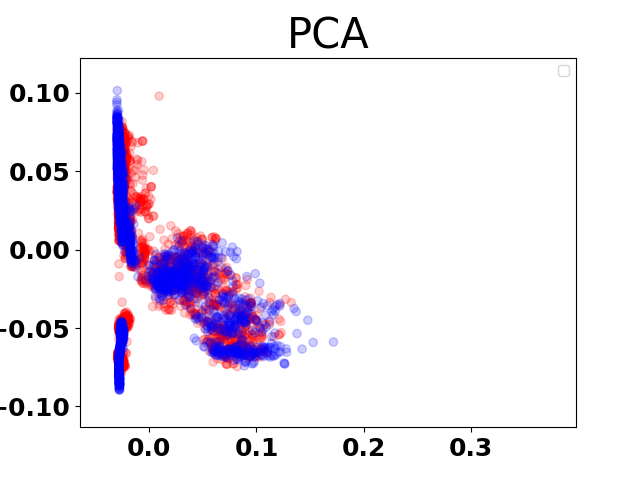}
\end{minipage}
}
\subfigure[VAE]{
\begin{minipage}[b]{.29\linewidth}
    \centering
    \includegraphics[width=1.1\linewidth]{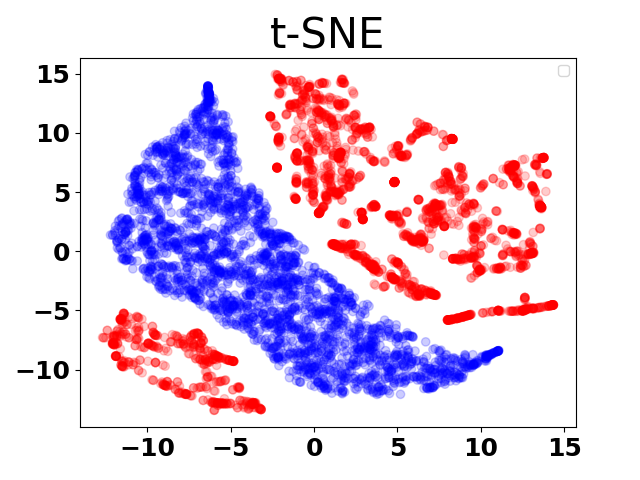} \\
    \includegraphics[width=1.1\linewidth]{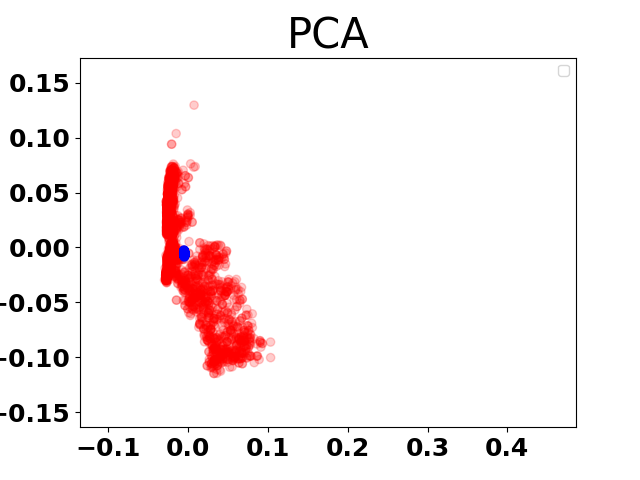}
\end{minipage}
}
\subfigure[WGAN]{
\begin{minipage}[b]{.29\linewidth}
    \centering
    \includegraphics[width=1.1\linewidth]{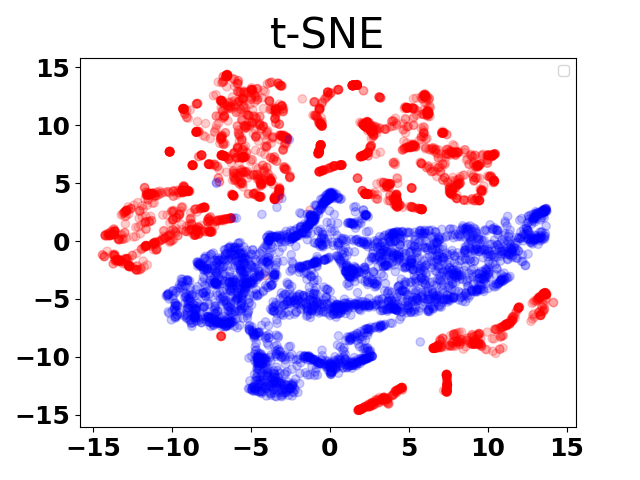} \\
    \includegraphics[width=1.1\linewidth]{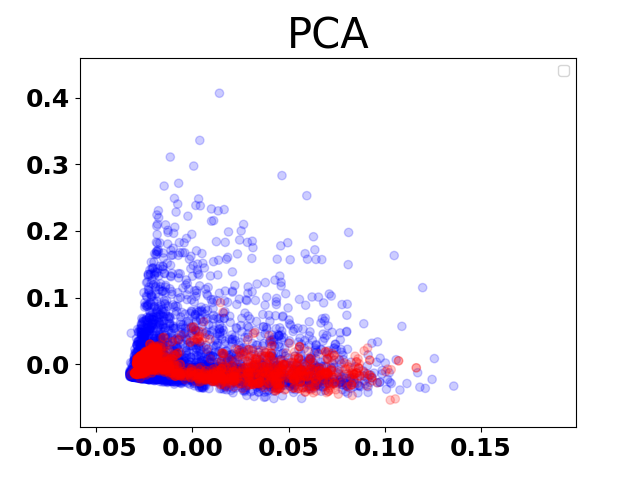}
\end{minipage}
}
\caption{The t-SNE and PCA plots of synthetic TMs in the GÉANT dataset: the red dots denote the true TM data, and the blue dots denote the synthetic TM data.}
\label{fig_4}
\end{figure}

\subsection{TM synthesis} \label{5d}

Fig.~\ref{fig_3} and Fig.~\ref{fig_4} present the t-SNE and PCA charts of TMs generated by our model and baseline models. In the figures, the red dots denote the true TM data, while the blue dots denote the synthetic TM data. From the figures, the synthetic TMs generated by our DDPM-based model showed markedly best overlap with true TM than other baselines. As the TMs in GÈANT dataset have much higher dimensions than the Abilene dataset, the synthetic TMs generated by VAE and WGAN in GÈANT suffer from a lower quality than in Abilene. However, our DDPM-based model can still produce high-quality TMs in GÈANT owing to the preprocessing module before the DDPM.

\subsection{TM Estimation} \label{5e}
\begin{figure}[htbp]
\centering
\subfigure[Abilene]{
\centering
\includegraphics[width=0.46\linewidth, height=0.35\linewidth]{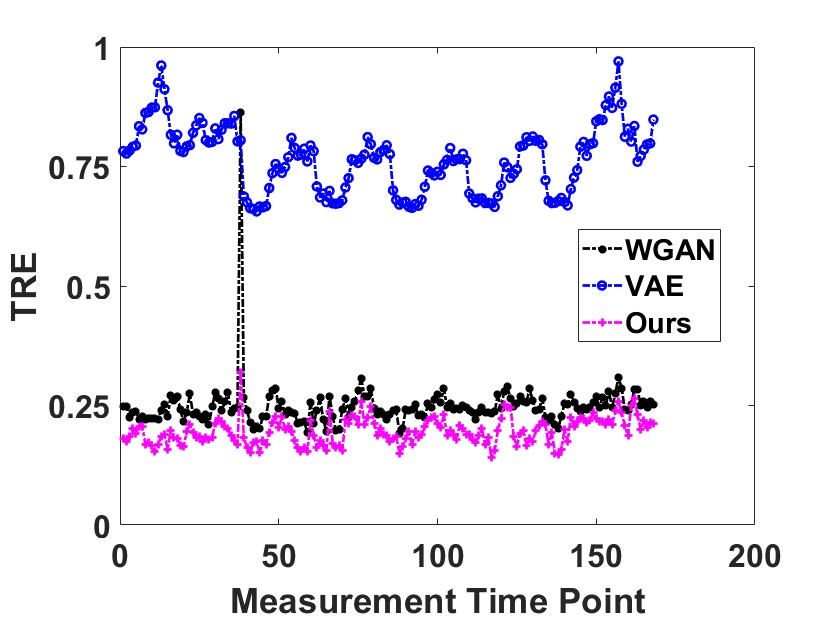}\label{fig_5a}
}
\subfigure[GÉANT]{
\centering
\includegraphics[width=0.46\linewidth, height=0.35\linewidth]{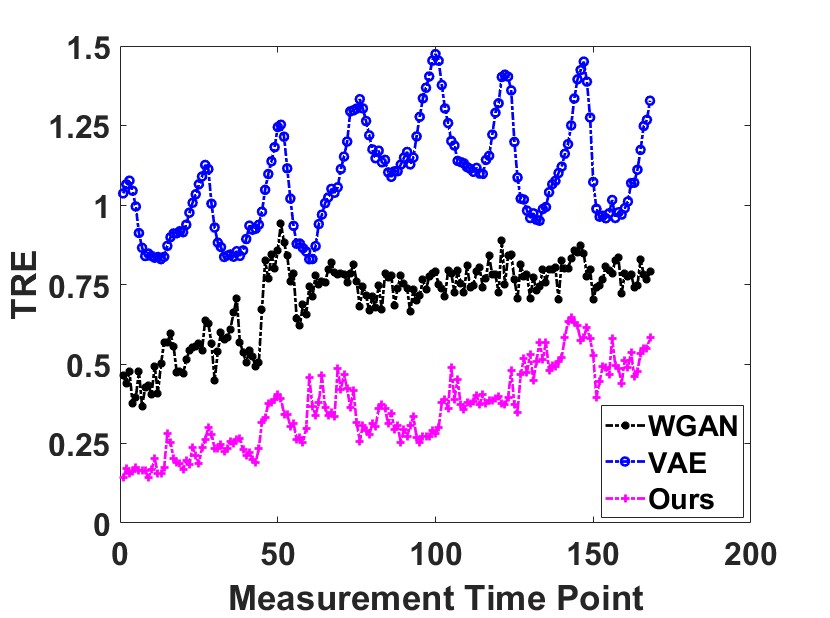}\label{fig_5b}
}
\caption{Temporal estimation errors.}
\label{fig_5}
\end{figure}

In this section, we evaluate the performances of our method and the baseline methods on TM estimation. Fig.~\ref{fig_5} plots the TREs of all three methods under two datasets. In both datasets, the $7$-days’ instances were aggregated into 168 records. Fig.~\ref{fig_5a} and Fig.~\ref{fig_5b} show that as the measurement time goes beyond, the curves of the three methods are gradually showing periodicity. Besides, DDPM and WGAN outperform VAE because they can better learn the structural information of training data. From Fig.~\ref{fig_5b}, in the GÈANT dataset, the accuracies of all methods gradually decrease with the measurement time going beyond due to the higher dimensional of TMs in GÈANT. Nevertheless, our method has an even more obvious superiority in the GÈANT dataset, owning to the pre-processing module in our model as well as the strong ability of DDPM on learning complex distributions. 

\begin{figure}[htbp]
\centering
\subfigure[Abilene]{
\centering
\includegraphics[width=0.46\linewidth, height=0.35\linewidth]{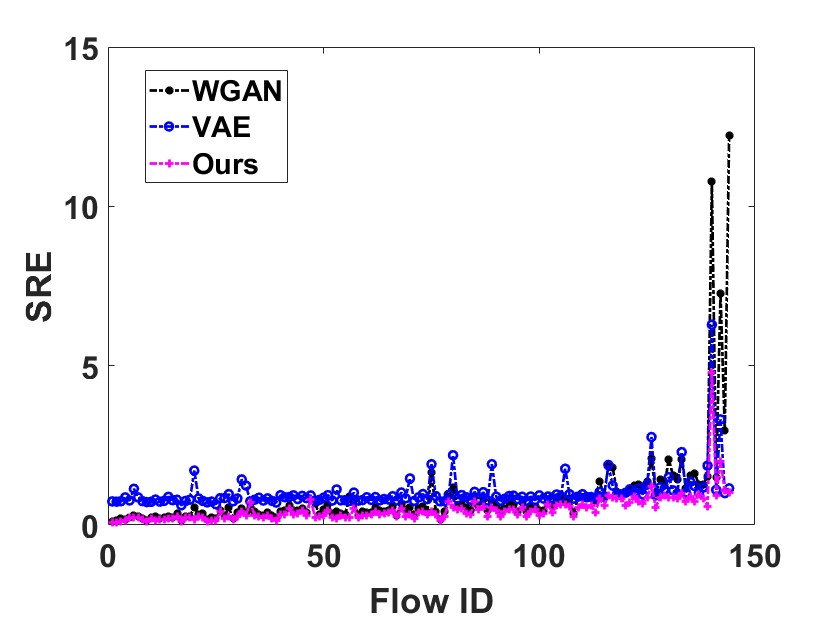}\label{fig_6a}
}
\subfigure[GÉANT]{
\centering
\includegraphics[width=0.46\linewidth, height=0.35\linewidth]{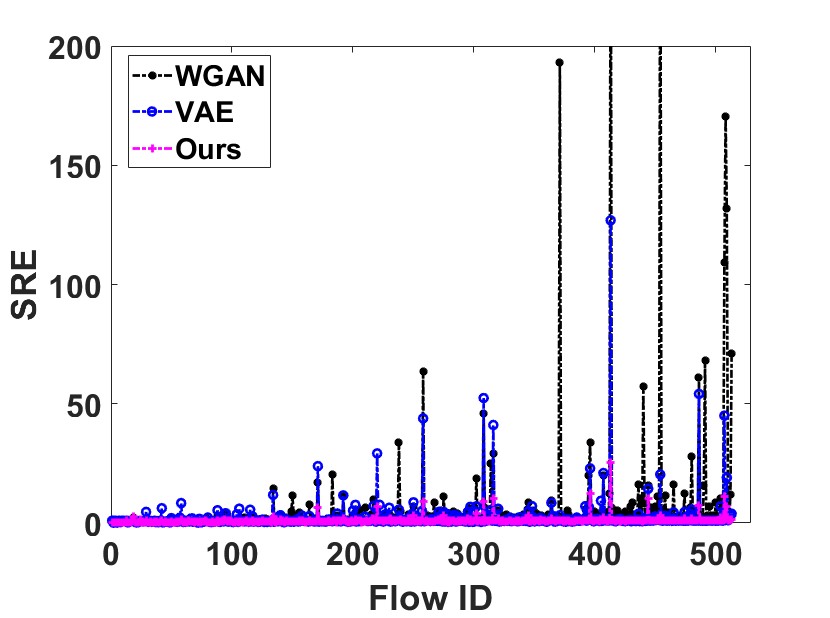}\label{fig_6b}
}
\caption{Spatial estimation errors.}
\label{fig_6}
\end{figure}

Fig.~\ref{fig_6} plots the SREs of each OD flow, where all the flows were sorted by their average volumes in ascending order. From the figures, it can be seen that with the volume of OD flows increasing in the network, SREs also present an increasing trend. That is because the large-size flows usually fluctuate within a greater range, leading to greater estimation errors. In addition, our DDPM-based method presents the most stable performance with the least estimation errors, while the baseline methods are more susceptible to volume changes. As shown in Fig.~\ref{fig_6b}, the phenomenon become obvious with higher dimensionality and sparsity. This indicates that our DDPM-based method is robust to the flow differences in TMs. 

\vspace{-0.8cm}

\vspace{-0.8cm}

\begin{figure}[htbp]
\centering
\subfigure[Abilene]{
\centering
\includegraphics[width=0.46\linewidth, height=0.35\linewidth]{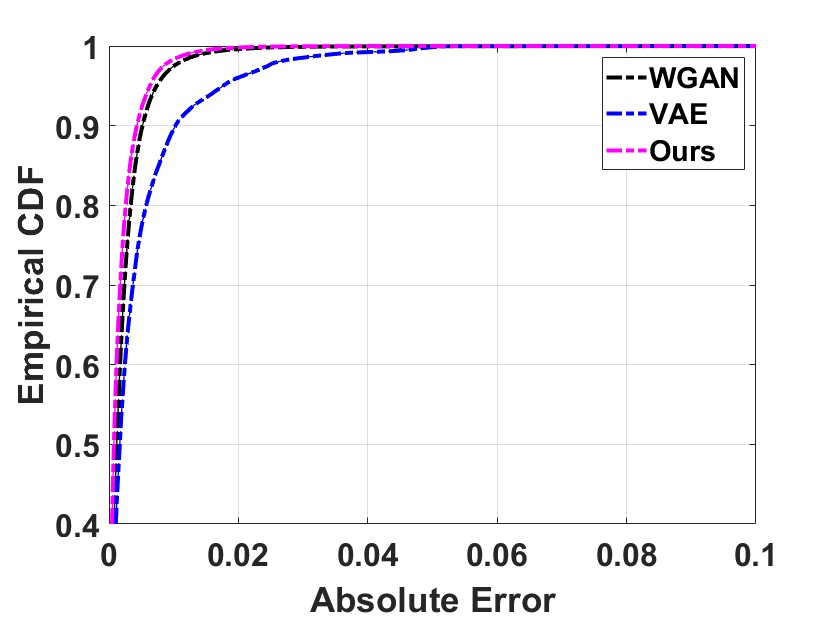}\label{fig_7a}
}
\subfigure[GÉANT]{
\centering
\includegraphics[width=0.46\linewidth, height=0.35\linewidth]{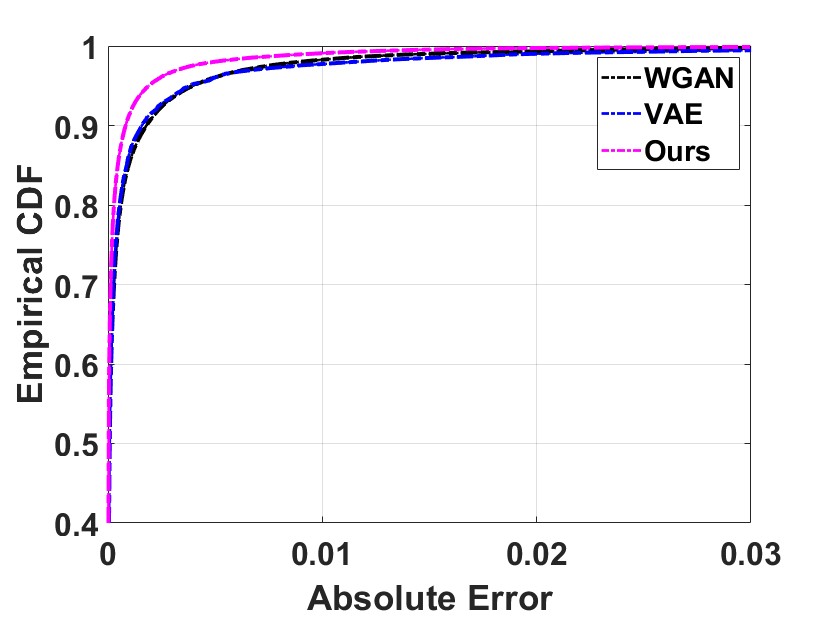}\label{fig_7b}
}
\caption{Empirical CDF of absolute errors.}
\label{fig_7}
\end{figure}

Fig.~\ref{fig_7} shows the respective cumulative distribution functions (CDFs) of absolute errors between the estimated flow volumes and the true flow volumes\footnote{All flow volumes have been divided by the maximum flow volume in each dataset}. From the two figures, our DDPM-based method has an evidently higher estimation accuracy, especially in the GÉANT dataset.

Table~\ref{tab2} and \ref{tab3} summarize the results for the three TME methods in two datasets. As shown in the two tables, our DDPM-based method outperforms the other two baselines on almost all metrics. Particularly, our method can guarantee much lower maximum errors in terms of all metrics in both datasets, which means the DDPM-based method has strong stability. Our DDPM-based method also has much lower mean errors and median errors among all metrics, which means the overall performance of our method significantly outperforms the other two baselines. Specifically, the DDPM-based method can improve the overall accuracy of the best baseline by $14\%$ in Abilene dataset, and $36\%$ in the GÉANT dataset.

\vspace{-0.2cm}
\begin{table}[htbp]
\caption{Estimation Errors of Abilene Dataset}
\begin{center}
\begin{tabular}{|c|c|c|c|c|}
\hline
\multicolumn{5}{|c|}{DDPM-based TME Method}\\
\hline
Metric& Mean& Median& Std& Max\\
\hline
RMSE ($\times 10^9$)& \textbf{0.0312}& \textbf{0.0290}& \textbf{0.0209}& \textbf{0.2880}\\ 
SRE& \textbf{0.4871}& \textbf{0.3797}& \textbf{0.4597}& \textbf{4.7922}\\
TRE& \textbf{0.1959}& \textbf{0.1927}& \textbf{0.0278}& \textbf{0.3182}\\
\hline
\multicolumn{5}{|c|}{WGAN-based TME Method}\\
\hline
Metric& Mean& Median& Std& Max\\
\hline
RMSE ($\times 10^9$)& 0.0362& 0.0335& 0.0234& 0.3240\\ 
SRE& 0.8728& 0.5327& 1.4640& 12.2259\\
TRE& 0.2479& 0.2426& 0.0337& 0.8631\\
\hline
\multicolumn{5}{|c|}{VAE-based TME Method}\\
\hline
Metric& Mean& Median& Std& Max\\
\hline
RMSE ($\times 10^9$)& 0.1464& 0.1385& 0.0390& 0.5716\\ 
SRE& 1.0236& 0.8778& 0.5824& 6.2839\\
TRE& 0.8412& 0.8459& 0.0456& 0.9771\\
\hline
\end{tabular}
\label{tab2}
\end{center}
\end{table}

\vspace{-0.5cm}
\begin{table}[htbp]
\caption{Estimation Errors of GÈANT Dataset}
\begin{center}
\begin{tabular}{|c|c|c|c|c|}
\hline
\multicolumn{5}{|c|}{DDPM-based TME Method}\\
\hline
Metric& Mean& Median& Std& Max\\
\hline
RMSE ($\times 10^7$)& \textbf{0.0255}& \textbf{0.0252}& 0.0078& \textbf{0.0432}\\ 
SRE& \textbf{1.0393}& \textbf{0.8484}& \textbf{1.6268}& \textbf{25.1756}\\
TRE& \textbf{0.3594}& \textbf{0.3623}& 0.1258& \textbf{0.6510}\\
\hline
\multicolumn{5}{|c|}{WGAN-based TME Method}\\
\hline
Metric& Mean& Median& Std& Max\\
\hline
RMSE ($\times 10^7$)& 0.0401& 0.0398& \textbf{0.0039}& 0.0486\\ 
SRE& 5.6521& 1.4715& 23.1754& 287.7491\\
TRE& 0.6795& 0.7239& 0.1231& 0.8868\\
\hline
\multicolumn{5}{|c|}{VAE-based TME Method}\\
\hline
Metric& Mean& Median& Std& Max\\
\hline
RMSE ($\times 10^7$)& 0.0630& 0.0626& 0.0126& 0.0897\\ 
SRE& 2.4148& 0.9681& 7.6698& 126.9252\\
TRE& 0.8791& 0.8742& \textbf{0.0853}& 1.1185\\
\hline
\end{tabular}
\label{tab3}
\end{center}
\end{table}
\vspace{-0.2cm}

\section{Conclusion} \label{6}
In this paper, we propose a novel DDPM-based TME method that can accurately estimate the TMs by inputting the link load measurements. DDPM, known as the state-of-the-art generative model, has a powerful ability on learning sample patterns and producing high-quality samples. To leverage DDPM to learn the traffic patterns effectively, we design a DDPM-based model which contains a preprocessing module to map the training TMs onto the embedding space, and a DDPM network to learn the distribution of TMs in the embedding space. To estimate the TMs from link load measurements, we transform the ill-posed inverse problem into a gradient-descent optimization problem. Specifically, we extract the impact factors in each diffusion step and optimize these factors together to produce an optimal TM that most conform with the input link loads. Finally, we compare the proposed method with recent benchmarks based on other generative models (i.e., VAE and GAN) using two real-world traffic datasets. The experimental results demonstrate that our method not only can produce the TMs that have the most similarity with the true TMs but also can output the most accurate estimations of TMs by inputting the link loads.


\bibliographystyle{IEEEtran}
\bibliography{main}

\end{document}